\documentclass[lettersize,journal]{IEEEtran}
\usepackage{amsmath,amsfonts}
\usepackage{algorithmic}
\usepackage{algorithm}
\usepackage{array}
\usepackage[caption=false,font=normalsize,labelfont=sf,textfont=sf]{subfig}
\usepackage{textcomp}
\usepackage{stfloats}
\usepackage{url}
\usepackage{verbatim}
\usepackage{graphicx}
\usepackage{cite}
\usepackage{csquotes}
\usepackage{rotating}
\usepackage{multicol}
\usepackage{multirow}
\usepackage{tabularray}
\usepackage{array}
\newcolumntype{C}[1]{>{\centering\arraybackslash}p{#1}}
\newcolumntype{L}[1]{>{\raggedright\arraybackslash}p{#1}}
\usepackage{times}
\usepackage{epsfig}
\usepackage{color}
\usepackage[normalem]{ulem}
\usepackage{booktabs}
\usepackage[colorlinks=true,linkcolor=black,citecolor=blue,urlcolor=blue,]{hyperref}
\usepackage{algorithm,algorithmic}

\usepackage[normalem]{ulem}

\begin{document}

\title{Generalised Co-Salient Object Detection}

\author{Jiawei~Liu, Jing~Zhang, Ruikai~Cui, Kaihao~Zhang, Weihao~Li, Nick~Barnes
\thanks{J, Liu, J, Zhang, R, Cui, K, Zhang, W, Li and N, Barnes are with the College of Engineering and Computer Science, The Australian National University, Canberra, ACT, Australia. E-mail: \{(Jiawei.Liu3, Jing.Zhang, Ruikai.Cui, Kaihao.Zhang, Weihao.Li, Nick.Barnes)@anu.edu.au\}}
}



\maketitle

\begin{abstract}
We propose a new setting that relaxes an assumption in the conventional Co-Salient Object Detection (CoSOD) setting by allowing the presence of \enquote{noisy images} which do not show the shared co-salient object. We call this new setting Generalised Co-Salient Object Detection (GCoSOD). We propose a novel random sampling based Generalised CoSOD Training (GCT) strategy to distill the awareness of inter-image absence of co-salient objects into CoSOD models. It employs a Diverse Sampling Self-Supervised Learning (D$\text{S}^{3}$L) that, in addition to the provided supervised co-salient label, introduces additional self-supervised labels for noisy images (being null, that no co-salient object is present). Further, the random sampling process inherent in GCT enables the generation of a high-quality uncertainty map highlighting potential false-positive predictions at instance level. To evaluate the performance of CoSOD models under the GCoSOD setting, we propose two new testing datasets, namely CoCA-Common and CoCA-Zero, where a common salient object is partially present in the former and completely absent in the latter. Extensive experiments demonstrate that our proposed method significantly improves the performance of CoSOD models in terms of the performance under the GCoSOD setting as well as the model calibration degrees.
\end{abstract}

\begin{IEEEkeywords}
Generalised Co-Salient Object Detection, Co-Salient Object Detection Testing Dataset, Uncertainty Estimation, Model Calibration.
\end{IEEEkeywords}

\section{Introduction}
Co-salient object detection (CoSOD) \cite{CoEGNet_CoSOD3k,GCAGC,CADC,DDM} aims to localise a common salient object in a group of images. Different from single image based salient object detection (SOD) \cite{EGNet,TEP,ITSD}, CoSOD performs group-wise salient object detection, and the resulting salient map of each image within the group should reflect the common salient attribute across the group. To this end, most of the existing CoSOD models focus on group-wise attention/correlation modeling, aiming to discover the common salient regions.

\begin{figure}[!htp]
\centering
\includegraphics[width=\linewidth]{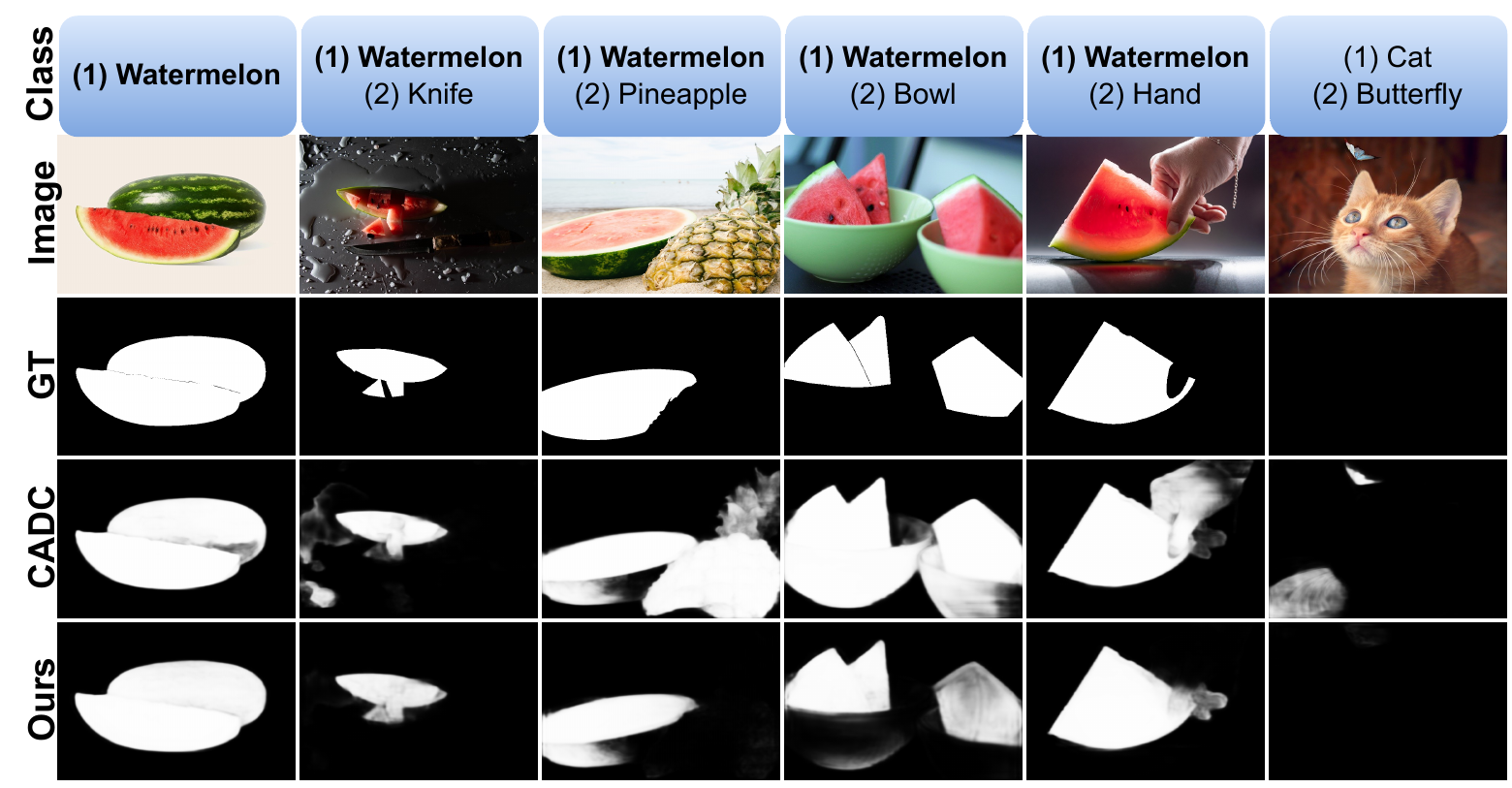}
\caption{
The group has \enquote{Watermelon} as the common salient object shared across its images except for an introduced \enquote{noisy image} which contains a \enquote{cat} and a \enquote{butterfly}. Existing models, such as CADC \cite{CADC}, find it challenging to acknowledge the absence of a common salient object in the \enquote{noisy image}. On the contrary, our proposed GCT, implemented on top of CADC, successfully avoids false-positive predictions in the \enquote{noisy image}.
}
\label{fig:critical_issues_cosod}
\end{figure}

We observe that the conventional CoSOD setting assumes the presence of co-salient object in each individual image belonging to the same pre-sorted group. The existing CoSOD models, despite performing well in localising the co-salient object across a group of pre-sorted images, find it challenging to deal with the unsorted image group, where co-salient object is only partially present. As illustrated in Fig.~\ref{fig:critical_issues_cosod}, CADC \cite{CADC}, one of the SOTA models in conventional CoSOD benchmarks, has difficulty in acknowledging the absence of the co-salient object in the introduced \enquote{noisy image} due to the knowledge distilled from the conventional CoSOD training. 

We propose a new setting that relaxes the conventional CoSOD setting by allowing the presence of varying numbers of \enquote{noisy images} that do not share the co-salient object of the pre-sorted group. We call this new setting as Generalised Co-Salient Object Detection (GCoSOD). It is motivated by a Photo Sorting problem that requires finding subgroups in photo collections that are each associated with a common topic/object without tags as prior knowledge. It is natural to assume that the entire photo collection does not have any common salient object that is universally present in each individual image and the presence ratio of common salient object in individual cases may vary. GCoSOD can be iteratively applied to the photo collections: (1) find the co-salient object of current iteration that has the most appearances in the collection; (2) relocate the retrieved images with the co-salient object to the corresponding subgroup.


To address the false-positive predictions of co-salient objects in the noisy images, we propose a generalised CoSOD training (GCT) strategy to distill the awareness of the potential absence of co-salient object in the inter-image level into CoSOD models. Instead of feeding the models with training images from the pre-sorted groups, GCT partially/wholly replaces them with images containing salient objects of distinct categories, regarded as \enquote{noisy images}. The \enquote{noisy images}, associated with complete-negative ground truth, can present at different levels ranging from 0 to 100\%, leading to two extreme cases: 1) 0\% of the images are noisy, which is the same as the conventional CoSOD setting, where co-salient object exists in each image of the group; and, 2) 100\% of the images are noisy, indicating that there is no overlap in terms of the salient object categories, or there is no common salient object. We find that GCT not only improves the handling of \enquote{noisy images} by existing CoSOD models, but also significantly improves their model calibration degrees \cite{on_calibration} (see Fig.~\ref{fig:reliability_diagrams_visualization} and Tab.~\ref{tab:calibration_comparison}).

The core of GCT lies in the Diverse Sampling Self-Supervised Learning (D$\text{S}^{3}$L) that randomly determines the noise level (the percentage of \enquote{noisy images}) and the categories of the \enquote{noisy images}. It associates each training image with diverse annotations: 1) with foreground areas highlighting the co-salient object when it exists; and, 2) with complete-negative ground truth when it is sampled as a \enquote{noisy image}. Further, the diverse groundtruth co-saliency maps enabled by D$\text{S}^{3}$L make it possible to generate high-quality uncertainty maps highlighting potential false-positive predictions at an instance level (see Fig.~\ref{fig:qualitative_ablation} and Fig.~\ref{fig:uncertainty_comparison}).


In addition to the training strategy, we also observe that existing CoSOD testing datasets pre-sort images into groups, where a unique co-salient object is present universally in each individual image within each pre-sorted group. Such groups are unsuitable to evaluate the ability of CoSOD models to handle \enquote{noisy images} in the GCoSOD setting. We re-arrange the CoCA dataset \cite{GICD_CoCA} and propose two new testing datasets, namely CoCA-Common and CoCA-Zero, where a common salient object is partially present in the former and completely absent in the latter. Extensive experiments demonstrate the superiority of our method in the GCoSOD setting where the \enquote{noisy images} are present to various extents.

We summarise our main contributions as: 1) we propose a GCT strategy to enable the existing CoSOD models to handle \enquote{noisy images} in our newly proposed GCoSOD setting; 2) we present a new uncertainty estimation technique via diverse groundtruth sampling, enabled by D$\text{S}^{3}$L;
3) we re-arrange the CoCA dataset\cite{GICD_CoCA} to produce two new CoSOD testing datasets, namely CoCA-Common and CoCA-Zero, to evaluate the performance of CoSOD models under the proposed GCoSOD setting where \enquote{noisy images} can present to various extents; 4) we perform extensive experiments to verify the effectiveness of the proposed GCT in improving existing CoSOD models in terms of \enquote{noisy image} handling and model calibration degree.

\section{Related Works}
\noindent\textbf{Co-Salient Object Detection:}
Traditional CoSOD methods are based on handcrafted features \cite{cao2014self,li2011co,li2014efficient,fu2013cluster}. These were succeeded by deep neural network (DNN) based approaches with more representative deep features. Early DNN based CoSOD models combine the representation power of convolutional neural networks (CNNs) with traditional methods \cite{zhang2016co}. Recent works explore the correlation of co-salient objects across the group of images with solely CNNs. They can be categorised into (1) affinity based methods \cite{GCoNet,li2019detecting,CADC,CoEGNet_CoSOD3k,CSMG,CoADNet,DeepACG,ICNet,DDM}, and (2) graph neural network (GNN) based methods \cite{wei2019deep,jiang2020co,GCAGC,jiang2019unified}.

Affinity based methods search for common features across a group of images based on their feature affinity to localise the co-salient objects. This has been achieved through pixel-level affinity \cite{GCoNet}, a recurrent co-attention unit \cite{li2019detecting}, self-attention mechanism \cite{CADC}, Principal Component Analysis \cite{CoEGNet_CoSOD3k,CSMG}, etc. CoADNet \cite{CoADNet} develops a two-stage aggregation model, utilising a combination of a spatial attention mechanism and channel attention mechanism, together with a distribution mechanism to find the co-salient object 
in different image groups. \cite{DeepACG} builds 4D correlation volumes on the Gromov-Wasserstein distance to discover co-salient object through pair-wise pixel similarities. In addition to common features, \cite{ICNet} further employs saliency priors to eliminate distractions from similar background where common objects can potentially appear. Graph Neural Networks (GNN) based methods construct graph models on deep features and find common features through graph clustering or graph propagation. \cite{wei2019deep} designs a graph manifold ranking strategy to propagate the group common features to localise co-salient object in individual images. \cite{jiang2020co,GCAGC} propose an adaptive graph to learn inter- and intra-image consistency. This idea is extended in \cite{jiang2019unified} by leveraging multiple graph models in a unified framework.

\begin{figure*}[!htb]
    \centering
    \includegraphics[width=\textwidth]{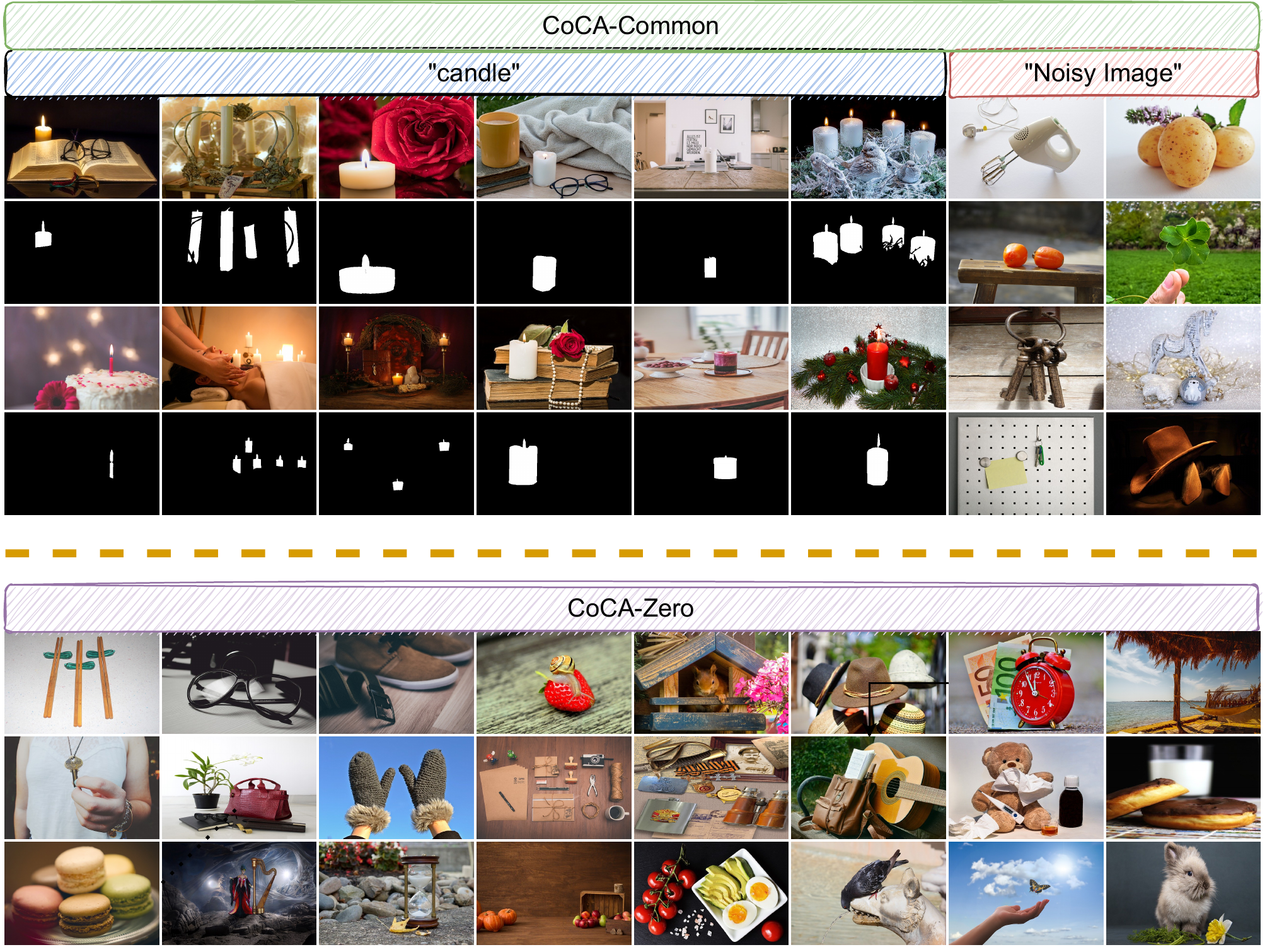}
    \caption{Samples from the proposed \enquote{CoCA-Common} dataset (top) and \enquote{CoCA-Zero} dataset (bottom). The \enquote{Candle} group of CoCA-Common has candle as the co-salient object shared across the primary images. It also includes noisy images whose salient objects have categories different from candle. On the other hand, the  group of CoCA-Zero consists of images with distinguished salient object categories from each other.}
\label{fig:two_new_cosod_dataset_visualization}
\end{figure*}

In this paper, we propose a GCT to distill the awareness of potential inter-image absence of the common salient object into the existing CoSOD models, improving their robustness to the presence of \enquote{noisy images} that do not share the group-wise co-salient object.

\noindent\textbf{CoSOD Datasets:}
Existing CoSOD testing datasets include MSRC \cite{MSRC}, iCoSeg \cite{iCoSeg}, CoSal2015 \cite{CoSal2015}, CoSOD3k \cite{CoEGNet_CoSOD3k} and CoCA \cite{GICD_CoCA}. MSRC and iCoSeg contains 233 images divided into 7 groups and 643 images divided into 38 groups respectively. High appearance consistencies are observed in both foreground salient objects and background scenes of those images, making them relatively easy for CoSOD. CoSal2015 \cite{CoSal2015} and CoSOD3k \cite{CoEGNet_CoSOD3k} are two large-scale CoSOD datasets, comprising of 2,015 and 3,316 images in 160 and 80 groups respectively. Despite the increase in size and variance in both foreground objects and background scenes, most of the images constitute single foreground objects, which make them easily solvable
with single RGB image based SOD models. This is addressed to some extent in the newly proposed CoCA \cite{GICD_CoCA} dataset where salient objects not belonging to the co-salient categories are introduced in each image as distractions.

\noindent\textbf{Model Calibration:} Guo et al. \cite{on_calibration} discover that modern deep neural networks are prone to making over-confident predictions. To remedy this issue, \cite{VAT_tpami,distribution_smooth} generate adversarial samples~\cite{Goodfellow2015ExplainingAH} to increase the diversity of training dataset. Kuamr et al. \cite{pmlr-v80-kumar18a} introduce a trainable model calibration error as a regularization term into their objective function.
\cite{can_you_trust_nips} investigates the effect of dataset shift on prediction accuracy and model calibration degree. Two main strategies have been widely studied to achieve model calibration, namely label relaxation \cite{rethink_inception} and temperature scaling \cite{on_calibration}. Label relaxation aims to relax the supervision signals, thus generating smoothed labels \cite{rethink_inception} or disturbed labels \cite{disturblabel}. ASLP \cite{liu2023model} proposes an adaptive stochastic label perturbation following the Maximum Entropy Inference \cite{jaynes1957information} of statistical mechanics. We propose a novel approach that augments training data with diversified self-supervised labels to reduce the overconfidence in falsely predicted co-salient objects.

\noindent\textbf{Uncertainty Estimation:}
Uncertainty estimation aims to evaluate model prediction in terms of its reliability \cite{kendall2017uncertainties}. Approaches have been explored in several dense segmentation tasks \cite{li2021uncertainty,mendel2020semi,TEP,liu2022modeling}. Existing techniques for uncertainty estimation usually adopt the Bayesian Neural Network \cite{izmailov2021bayesian,ober2021global} or stochastic prediction network \cite{depeweg2018decomposition,kwon2020uncertainty}. The former focus on model uncertainty, indicating inadequate unawareness of the true model that generates the observed data, and latter is usually designed to model data uncertainty, representing the inherent labeling noise. In this paper, we introduce a diverse sampling based uncertainty estimation method that is capable of highlighting potential false-positive predictions at an instance level.

\section{Our Method}
\subsection{Overview}
For co-salient object detection, we have a dataset $D = \{G_{k}\}_{k=1}^{K}$, consisting of $K$ groups of images with the corresponding ground truth saliency maps, which is $G_{k} = \{x_{kj}, y_{kj}\}_{j=1}^{N_{k}}$, where $x_{kj}$ and $y_{kj}$ indicate an image-groundtruth pair of index $j$ from the $k^{th}$ group, and $N_{k}$ is the size of group $G_{k}$. Given the training dataset $D$, the typical training procedure of existing CoSOD models performs co-salient object detection within each pre-sorted group, where a common salient object is present in each individual image of the same group. 
Despite achieving strong performance under the conventional CoSOD setting, existing CoSOD models are not robust to the introduction of \enquote{noisy images} which do not share the common salient object.

To avoid false-positive predictions for the \enquote{noisy images}, we propose a Generalised CoSOD Training strategy that allows the model to detect the absence of common salient objects in \enquote{noisy images}. In addition, to evaluate the performance of CoSOD models under the proposed GCoSOD setting, we introduce two new testing datasets, CoCA-Common and CoCA-Zero, which are generated by re-arranging the CoCA \cite{GICD_CoCA} testing dataset, with samples presented in Fig.~\ref{fig:two_new_cosod_dataset_visualization}. The former contains groups where only a fraction of the images contain the common salient object, and the latter contains groups of images with no co-salient object at all.

\subsection{Generalised CoSOD Training Strategy}
We propose a random sampling based Generalised CoSOD Training strategy, 
that enables CoSOD models to deal with scenarios where there is complete or partial absence of the common salient object across the images of a group. Firstly, instead of requiring a common salient object to be present in each individual image of a group, we relax this requirement by defining the co-salient object as a salient object with the most appearances in the images of the same group. The proposed GCT randomly introduces \enquote{noisy images} at various ratios. 
It further employs a Diverse Sampling Self-Supervised Learning (D$\text{S}^{3}$L) to introduce additional self-supervised labels, with complete \textbf{0} indicating full background, for images drawn as \enquote{noisy images} with no common salient object.
With this, for each pre-sorted co-salient group, we define it as a \enquote{primary group}, associated with a corresponding \enquote{secondary group} containing images that do not share the co-salient object.

\begin{figure*}[htb!]
    \centering
    \includegraphics[width=\linewidth]{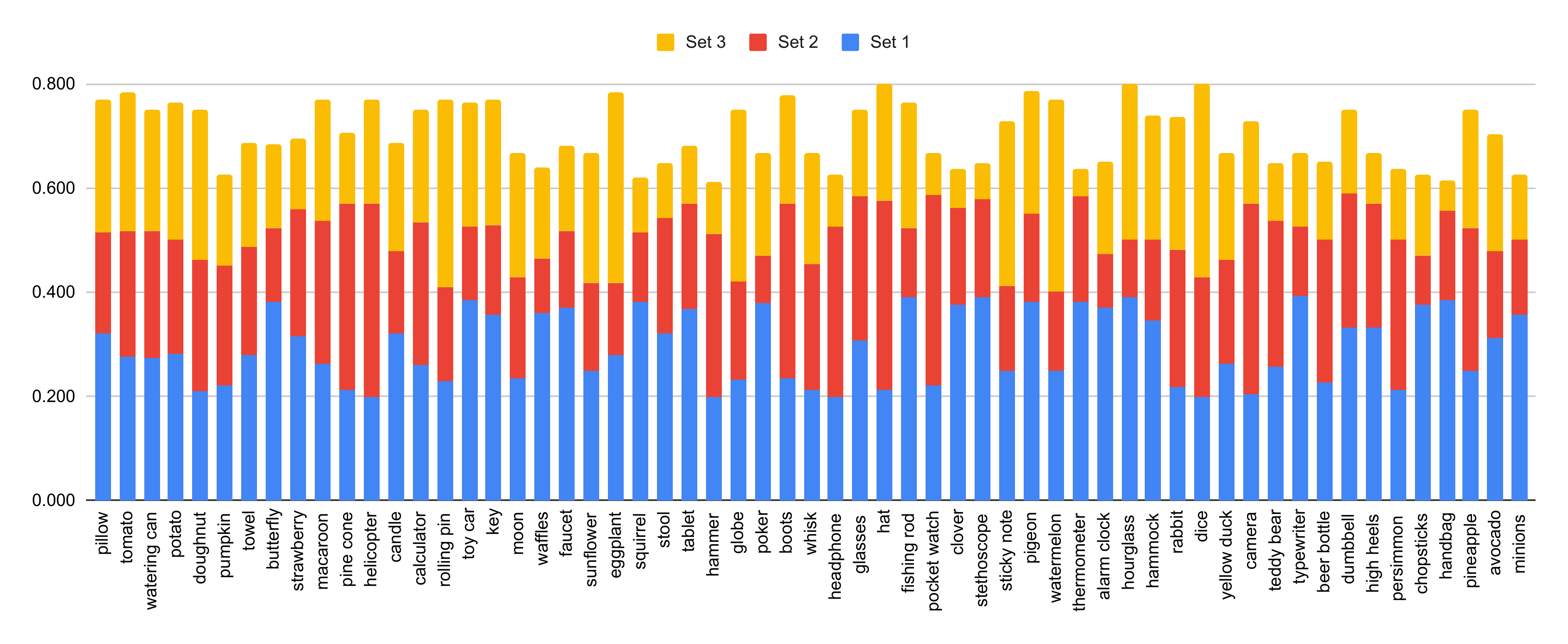}
    \caption{Distribution of primary group ratio in CoCA-Common dataset. Each category appears three times with the primary group ratio in ranges [0,2, 0.4), [0.4, 0.6) and [0.6, 0.8] respectively.}
    \label{fig:dataset_distributions}
\end{figure*}

\noindent\textbf{Diverse Sampling Self-Supervised Learning (D$\text{S}^{3}$L):}
We randomly draw images from the secondary groups as \enquote{noisy images} to partially replace images in the primary group. 
The noisy-image-replacement process: (1) randomly determines the replacement ratio by sampling from a uniform distribution as: $\tilde{r} \sim U(0, 1)$, and computes the number of replacement images as $r = \lfloor N_{i} \times \tilde{r} \rfloor$, where $\lfloor \cdot \rfloor$ is a floor function; (2) We randomly select $r$ groups from the $K - 1$ secondary groups and draw a single noisy image from each of them to replace the images in the primary group. The stochastic attribute imbued in the random selection of noisy image distills the awareness of potential absence of co-salient object in some noisy inputs, which could contain non-co-salient object of various classes, textures and shapes.


\noindent\textbf{Extreme Cases:} 
Two extreme cases exist in our random sampling based training dataset preparation process: 1) the primary group is unchanged with proportion of the secondary groups equal to 0; and 2) the primary group is totally replaced with secondary groups, where no common salient objects exists across the new group. With the first case, our training dataset is the same as the conventional CoSOD setting, where common salient objects exist across the group of images. The second case is designed to expose the model to scenarios where the proportion of images containing the common salient object is below certain threshold. 
For most of the cases, our random sampling strategy generates new groups with only a subset of images containing no common salient objects, which is consistent with real-life scenarios (i.e., an imperfectly curated set).

\subsection{Uncertainty Estimation via Random Sampling}
\label{sec:random_sampling_for_uncertainty_estimation}
The \enquote{random sampling} process inherent to the Generalised CoSOD Training creates multiple annotations for each training image: 1) when the image belongs to the primary group, its ground truth includes foreground areas covering the co-salient object; 2) when the image belongs to the secondary group, its ground truth is completely negative, indicating the absence of the co-salient object. 
The divergence modelling, conditioned on the group context, improves the model awareness of the conditional occurrence of a co-salient object. It enables the model to produce less confident predictions in difficult cases where the noisy salient object demonstrates certain appearance similarity to the common salient object. We define the uncertainty as the entropy of the prediction:
\begin{equation}
\begin{aligned}
u_{ki} &= \mathbb{H}(p_{\theta}(\hat{y}_{ki} | \{x_{kj}\}_{j=1}^{N_{k}})) \\
&= - p_{\theta}(\hat{y}_{ki} | \{x_{kj}\}_{j=1}^{N_{k}}) \log (p_{\theta}(\hat{y}_{ki} | \{x_{kj}\}_{j=1}^{N_{k}}) + \epsilon) 
\end{aligned}
\label{eqn:uncertainty_estimation_entropy}
\end{equation}
where $ p_{\theta}(\hat{y}_{ki} | \{x_{kj}\}_{j=1}^{N_{k}}) \in \mathbb{R}^{H \times W}$ is a normalised conditional co-saliency prediction with $H$ and $W$ representing the original image sizes, $i$ and $j$ indexes the samples in group $k$, and $k$ indexes the co-saliency groups, $\epsilon = 10^{-6}$ is introduced to avoid the $\log(0)$ error.

\subsection{New Testing Datasets for GCoSOD Setting}
We propose two new CoSOD testing datasets, namely \enquote{CoSOD-Common} and \enquote{CoSOD-Zero}, to measure the ability of CoSOD models to handle \enquote{noisy images} that do not share the common salient object. These two datasets are built by re-arranging the CoCA \cite{GICD_CoCA} dataset.

\subsubsection{CoCA-Common} CoCA-Common contains 177 groups of images where each group consist of images only some of which show a co-salient object. There are 59 distinct groups, each having three different co-salient object presence ratios falling in the ranges of [0.2, 0.4), [0.4, 0.6) and [0.6, 0.8] respectively. The overall presence ratio of co-salient objects is between 0.2 and 0.8. 
Fig.~\ref{fig:dataset_distributions} presents the co-salient object presence ratios of CoCA-Common. The CoCA-Common dataset is designed to measure the robustness of CoSOD models in scenarios where \enquote{noisy images} are present in different scales. 

\subsubsection{CoCA-Zero} CoCA-Zero contains 55 groups of images, where each group may have a different number of images, and there is no common salient object across the images of each group. These groups are formed by selecting at most one image per group from the CoCA \cite{GICD_CoCA} dataset. Visual examination is applied as post-processing, ensuring no accidental occurrence of co-salient object in each group. For example, the \enquote{\textit{Person}} category has a 78\% presence in the \enquote{\textit{Basketball}} group of CoCA \cite{GICD_CoCA}, making it an unwanted, yet commonly presented secondary salient object. CoCA-Zero is designed to measure the model behaviour in the scenarios where no common salient object exists.

\subsection{Evaluation} 
With the two new testing datasets, we aim to produce $\textbf{0}$ (full background) for images containing no co-salient object. We find that widely used CoSOD evaluation metrics (F-measure $F_\beta$, S-measure \cite{S_measure} $S$ and E-measure \cite{E_measure} $\xi$) are unable to evaluate CoSOD models under the GCoSOD setting. For an image with $\textbf{0}$ as ground truth, as the true positive is 0, we then have $\text{precision}=0$, and the resulting F-measure $F_\beta=(1+\beta^2).\frac{\text{precision}.\text{recall}}{\beta^2 \text{precision} + \text{recall}}$ is also 0, which is independent of the quality of the predictions. Structure measure (S-measure) \cite{S_measure} evaluates the structure similarity between the prediction and ground truth in both region and object levels. Similar to the F-measure, based on the region and object levels mean foreground predictions $\bar{s}$, the two corresponding structure similarities ($S_r$ and $S_o$ for region and object level similarity respectively) are $\textbf{0}$ for all predictions, making S-measure ineffective in measuring model prediction in our scenario. The Enhanced-alignment measure (E-measure) \cite{E_measure} computes a bias matrix for the prediction ($\varphi_{s}$) and the ground truth map ($\varphi_{y}$) respectively, which is the distance between each pixel within the prediction or the ground truth and its global mean. The final alignment matrix $\xi$ is then defined as $\xi = \frac{2*\varphi_{s}\circ\varphi_{y}}{\varphi_{s}\circ\varphi_{s}+\varphi_{y}\circ\varphi_{y}}$,
where $\circ$ is the Hadamard product. Given accurate predictions with $\textbf{0}$ as ground truth, we have $\varphi_{s}=0$, leading to $\xi=0$ in this case.

Considering that the conventional metrics are less effective for evaluating the model performance under the proposed GCoSOD setting, we adopt the widely used mean Intersection-over-Union (mIoU) metrics. For our case with $\textbf{0}$ as ground truth, the resulting $\text{mIoU}=1$ for accurate predictions is consistent with the quality of the prediction.

\section{Experimental Results}
\subsection{Settings}

\subsubsection{Implementation Details}
We adopt CADC \cite{CADC} as our baseline to implement the proposed GCT (GCT-CADC). The model is initialised with ImageNet pretrained weights for the VGG-16 backbone. We keep other settings of the original work unchanged by setting the training size to 256 $\times$ 256 and batch size to 14. 
For data augmentation, we employ random flipping and random cropping. The model is trained with the Stochastic Gradient Descent (SGD) optimiser for 60,000 iterations. The initial learning rate is set to 0.01 and then halved at the $\text{20,000}^{\text{th}}$ and the $\text{40,000}^{\text{th}}$ iteration. 
We perform all ablation study experiments on GCT-CADC and interchangeably term it as \enquote{Ours}. We also generalise our proposed GCT onto ICNet \cite{ICNet} and DCFM \cite{DDM} to produce GCT-ICNet and GCT-DCFM and show the results in Tab.~\ref{tab:benchmark_CoSOD_Common_and_Zero}.

\subsubsection{Datasets}
Following the baseline model \cite{CADC}, we use a subset of COCO dataset \cite{COCO}, containing 9,213 images in 65 groups, and a synthetic dataset \cite{CADC} for training. We use our proposed datasets, CoSOD-Common and CoSOD-Zero, to test the generalisation ability of existing CoSOD models. Further, we use our proposed datasets, CoCA-Common and CoCA-Zero, to test the performance of the existing CoSOD models in the GCoSOD setting, and three widely used CoSOD datasets, including CoSal2015 \cite{CoSal2015}, CoSOD3k \cite{CoEGNet_CoSOD3k} and CoCA \cite{GICD_CoCA}, to evaluate our method under the conventional CoSOD setting.


\begin{figure*}[htb!]
\centering
\includegraphics[width=0.95\textwidth]{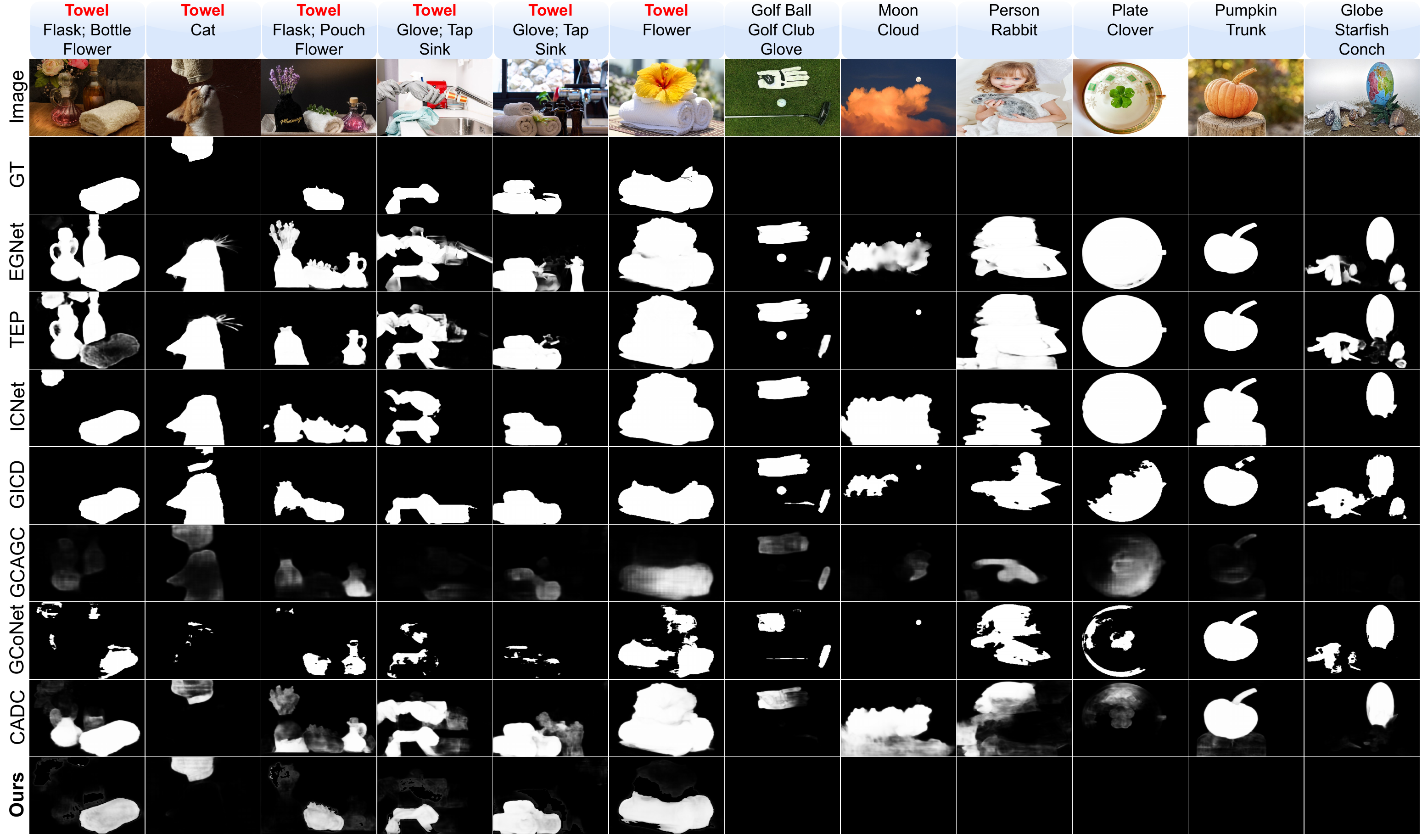} 
\vfill
\includegraphics[width=0.95\textwidth]{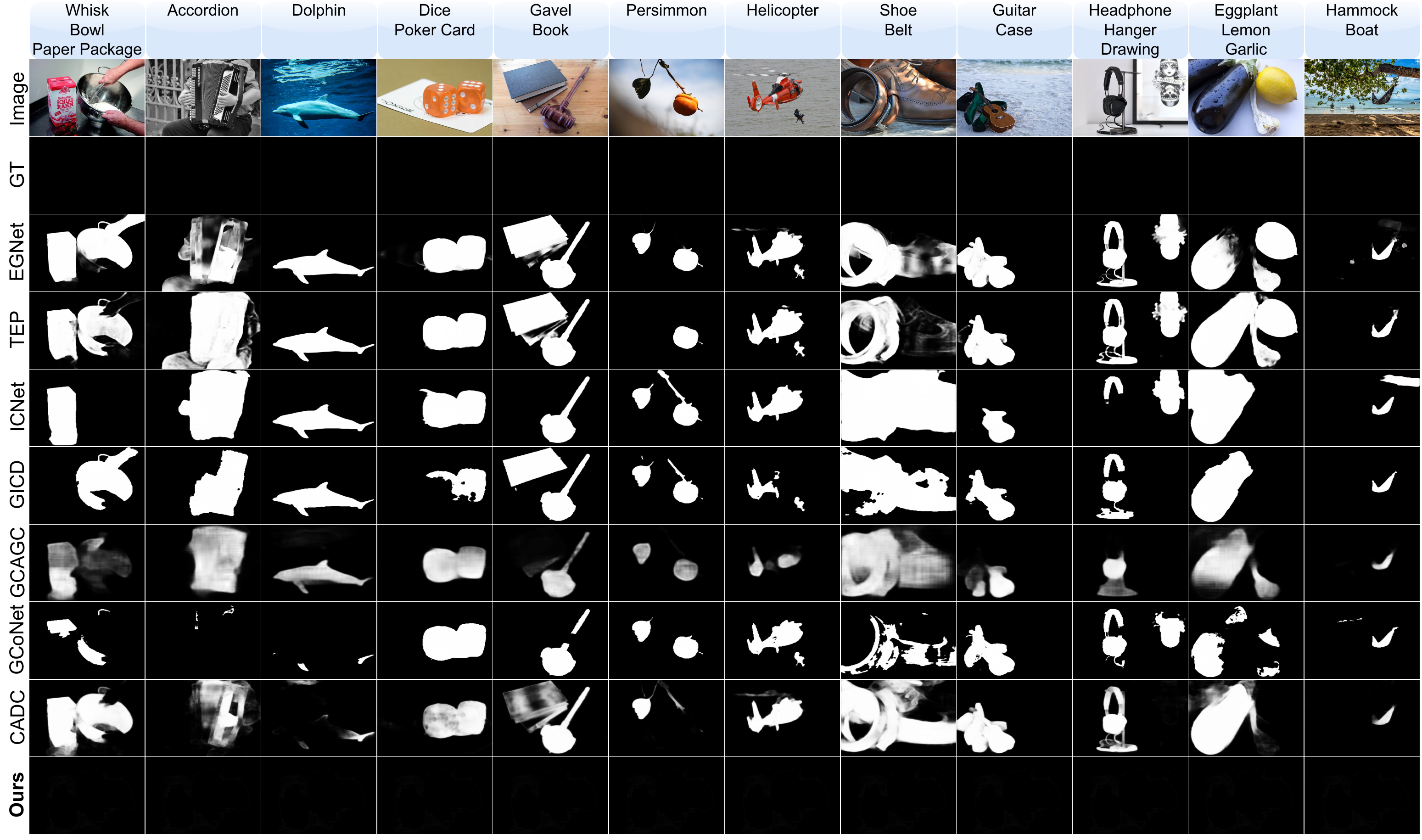}
\caption{Qualitative results of existing SOD and CoSOD methods and our model on CoCA-Common (top) and CoCA-Zero (bottom) datasets. The displayed images in each case belong to the same group. Images that do not contain the co-salient object are associated with complete-negative groundtruth).}
\label{fig:qualitative_results}
\end{figure*}

\subsubsection{Co-saliency Metrics}
We adopt the mean Intersection-over-Union (mIoU) and Mean Absolute Error (MAE) to evaluate the performance in GCoSOD setting, and the maximum F-measure ($F_{\beta}^{\mathrm{max}}$), Structure-measure ($S_{\alpha}$) \cite{S_measure} and maximum E-measure ($E_{\xi}^{\mathrm{max}}$) \cite{E_measure} and MAE for evaluation in the conventional CoSOD setting.

\subsubsection{Calibration Metrics}
We adopt two calibration measures to measure the model calibration degree 
including Expected Calibration Error (ECE)~\cite{ece_measure} and reliability diagrams~\cite{Degroot1983TheCA,predicting_good_probability}. We adopt the equal-width ECE \cite{guo2017calibration} defined as:
\begin{equation}
    \text{ECE} = \sum_{i=1}^{K} \frac{|B_{i}|}{N} |acc(B_{i}) - conf(B_{i}))
\end{equation}
where predictions are grouped into $K$ bins and $acc(B_{i})$ and $conf(B_{i})$ represent mean accuracy and mean confidence of predictions belonging to the $i^{th}$ bin, $|B_{i}|$ is the size of the $i^{th}$ bin and $N$ is the total number of predictions. Here we set the number of bins to $K = 10$. Reliability diagram plots the bin-wise accuracy against the oracle which has perfectly balanced bin-wise prediction accuracy and confidence.

\begin{table}[htb!]
\centering
\renewcommand{\arraystretch}{1.3}
\renewcommand{\tabcolsep}{1.0mm}
\caption{CoCA-Common and CoCA-Zero benchmark.}
\begin{tabular}{L{0.12\linewidth}L{0.18\linewidth}|C{0.12\linewidth}C{0.12\linewidth}|C{0.12\linewidth}C{0.12\linewidth}}
    \toprule
    \multicolumn{2}{c}{\multirow{2}{*}{Method}} & \multicolumn{2}{|c|}{CoCA-Common} & \multicolumn{2}{c}{CoCA-Zero} \\
    & & mIoU $\uparrow$ & $\mathcal{M}\downarrow$ & 
    mIoU $\uparrow$ & $\mathcal{M}\downarrow$ \\
    \midrule
    \multirow{2}{*}{\parbox{1.2cm}{SOD\\ Methods}} & EGNet \cite{EGNet} & 48.72 & 0.205 & 37.84 & 0.244\\
    & TEP \cite{TEP} & 48.41 & 0.211 & 37.60 & 0.248\\
    \midrule
    \multirow{5}{*}{\parbox{1.2cm}{CoSOD Methods}} & ICNet \cite{ICNet} & 51.37 & 0.191 & 38.12 & 0.245\\
    & GICD \cite{GICD_CoCA} & 52.99 & 0.150 & 40.90 & 0.190\\
    & GCAGC \cite{GCAGC} & 62.25 & 0.077 & 67.80 & 0.054\\
    & GCoNet \cite{GCoNet} & 56.22 & 0.093 & 46.92 & 0.087\\
    & CADC \cite{CADC} & 53.82 & 0.146 & 42.75 & 0.172\\
    & DCFM \cite{DDM} & 58.29 & 0.087 & 48.45 & 0.095\\
    \midrule
    \multirow{3}{*}{\parbox{1.2cm}{GCoSOD Methods}} 
    & GCT-ICNet & \textbf{63.70} & 0.084 & \textbf{83.36} & \textbf{0.024}\\
    & GCT-DCFM & \textbf{75.01} & \textbf{0.070} & \textbf{91.46} & \textbf{0.012}\\
    & GCT-CADC & \textbf{72.22} & \textbf{0.075} & \textbf{89.60} & \textbf{0.017}\\
    \bottomrule
\end{tabular}
\label{tab:benchmark_CoSOD_Common_and_Zero}
\end{table}

\subsection{Performance Under the GCoSOD Setting}
\subsubsection{Quantitative Comparison}
Tab.~\ref{tab:benchmark_CoSOD_Common_and_Zero} shows the performances of the SOTA CoSOD and SOD methods on the proposed CoCA-Common and CoCA-Zero datasets. Most of the methods achieve mIoU of less than 50\% on the CoCA-Zero dataset and less than 60\% on the CoCA-Common dataset. Without inter-image relationship modeling, the SOD methods tend to find all salient objects rather than the co-salient ones, leading to serious over-segmentation and poor mIoU and MAE scores. However, it is worth noting that SOD models are not trailing far behind some of their CoSOD counterparts. For example, EGNet \cite{EGNet} and TEP \cite{TEP} achieve similar performance to ICNet \cite{ICNet}, which has a strong reliance on saliency priors. With the extra co-saliency modeling, CoSOD models achieve relatively better performance than SOD models on the two new CoSOD testing datasets. Benefitting from its small inference size, GCAGC \cite{GCAGC} achieves the best performance among existing techniques. Instead of feeding the entire group of images to the model, GCAGC \cite{GCAGC} randomly selects five images from the group for co-salient map generation. Images within the small group (five images in GCAGC \cite{GCAGC}) are more likely to have a high variance in terms of the appearance and the salient object category, enabling the model to predict the absence of a co-salient object. Taking the entire group of images as input, our method significantly improved over the baselines (CADC \cite{CADC}, ICNet \cite{ICNet} and DCFM \cite{DDM}) and outperforms all existing CoSOD and SOD methods on both CoCA-Common and CoCA-Zero datasets.


\subsubsection{Qualitative Comparison}
Fig.~\ref{fig:qualitative_results} shows the predictions of existing CoSOD models and GCT-CADC on the CoCA-Common dataset. It illustrates that existing CoSOD models are not robust to the introduction of \enquote{noisy images} to a group of images sharing a common salient object, and they are prone to mistakenly predicting salient objects from the secondary groups as the common salient object. GCAGC \cite{GCAGC}, though outperforming other CoSOD models on the \enquote{noisy images}, tends to predict the absence of co-salient objects regardless of their true existence as a result of its small group size during inference. On the contrary, GCT-CADC is robust to both intra-image and inter-image noisy salient objects while accurately segmenting the common salient objects if present. 

\begin{table}[tbh!]
    \centering
    \scriptsize
    \renewcommand{\arraystretch}{1.2}
    \renewcommand{\tabcolsep}{1.2mm}
    \caption{$ECE$ of existing techniques and ours on benchmark co-saliency testing datasets our proposed testing datasets.}
    \begin{tabular}{l|ccccc}
        \toprule
        Method & CoCA & CoSOD3k  & CoSal2015 & CoCA-Common& CoCA-Zero \\
        \midrule
        ICNet \cite{ICNet} & 0.143 & 0.086 & 0.051 & 0.187 & 0.242 \\
        GICD \cite{GICD_CoCA} & 0.123 & 0.071 & 0.063 & 0.148 & 0.187 \\
        GCAGC \cite{GCAGC} & \textbf{0.048} & \textbf{0.036} & 0.032 & 0.043 & 0.020 \\
        GCoNet \cite{GCoNet} & 0.101 & 0.067 & 0.060 & 0.090 & 0.083 \\
        CADC \cite{CADC} & 0.086 & 0.042 & 0.021 & 0.102 & 0.132 \\
        DCFM \cite{DDM} & 0.082 & 0.060 & 0.060 & 0.085 & 0.092 \\ \hline
        GCT-ICNet & 0.080 & 0.052 & 0.026 & 0.051 & 0.016 \\
        GCT-DCFM & 0.062 & 0.038 & 0.020 & 0.030 & 0.017\\
        GCT-CADC & 0.058 & \textbf{0.036} & \textbf{0.016} & \textbf{0.022} & \textbf{0.007}\\ 
        \bottomrule
    \end{tabular}
    \label{tab:calibration_comparison}
\end{table}


\begin{figure*}[htb!]
\centering
\includegraphics[width=\textwidth]{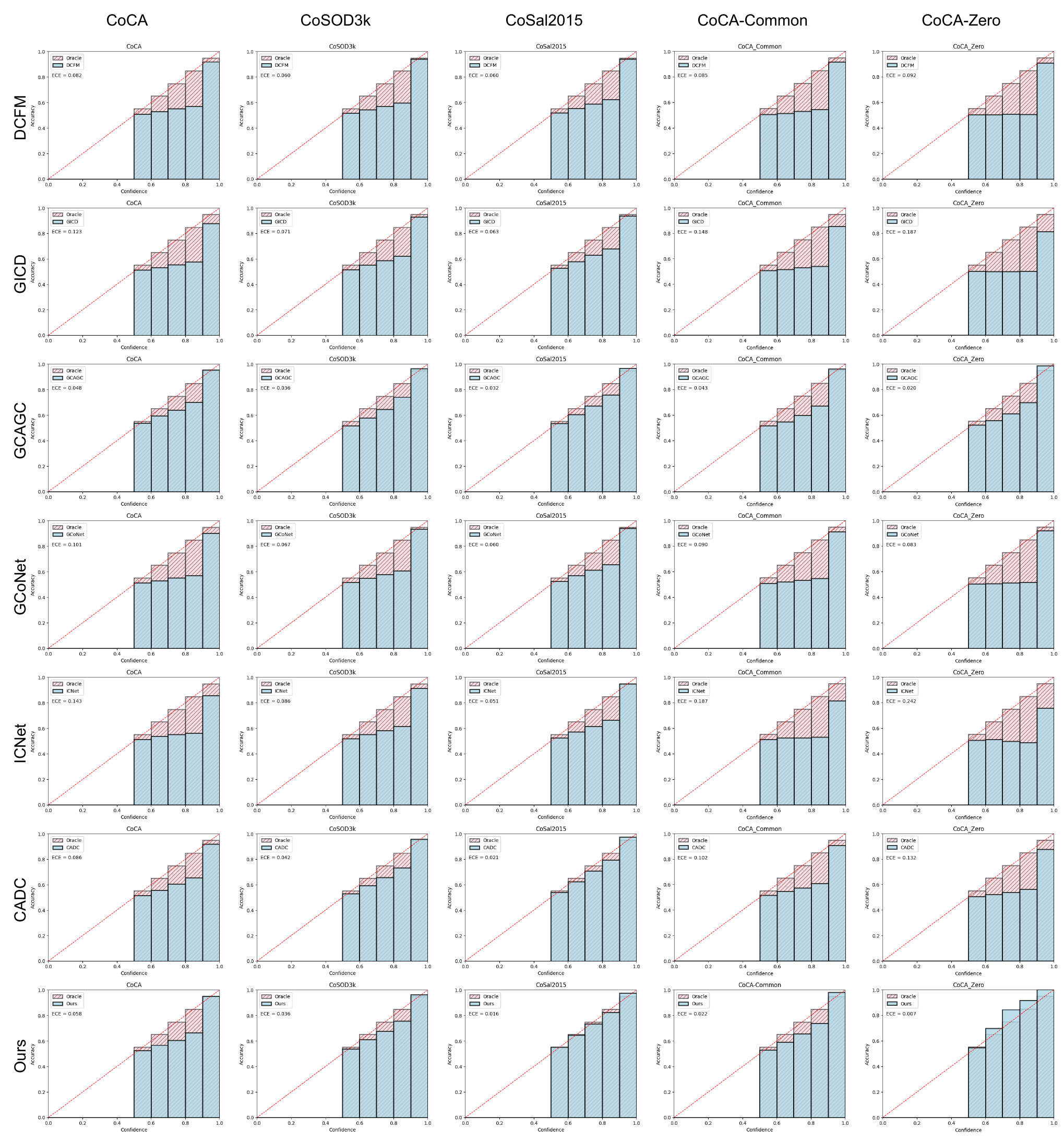}
\caption{Expected Calibration Error of 6 existing CoSOD models \cite{GICD_CoCA,CADC,DDM,GCoNet,ICNet} and our method (GCT-CADC) on both the conventional CoSOD testing datasets (CoCA \cite{GICD_CoCA}, CoSOD3k \cite{CoEGNet_CoSOD3k}, CoSal2015\cite{CoSal2015}) and the generalised CoSOD testing datasets (CoCA-Common and CoCA-Zero).}
\label{fig:reliability_diagrams_visualization}
\end{figure*}

\begin{table*}[tbh!]
    \centering
    \renewcommand{\arraystretch}{1.3}
    \renewcommand{\tabcolsep}{1.0mm}
    \caption{Performances of model with the proposed GCT strategy on conventional CoSOD testing datasets.}
    \begin{tabular}{L{0.10\textwidth}|C{0.05\textwidth}C{0.05\textwidth}C{0.05\textwidth}C{0.05\textwidth}|C{0.05\textwidth}C{0.05\textwidth}C{0.05\textwidth}C{0.05\textwidth}|C{0.05\textwidth}C{0.05\textwidth}C{0.05\textwidth}C{0.05\textwidth}}
        \toprule
        \multirow{2}{*}{Method} & \multicolumn{4}{c}{CoCA \cite{GICD_CoCA}} & \multicolumn{4}{|c}{CoSOD3k \cite{CoEGNet_CoSOD3k}} & \multicolumn{4}{|c}{CoSal2015 \cite{CoSal2015}} \\
        & F $\uparrow$ & E $\uparrow$  & S $\uparrow$ & $\mathcal{M}\downarrow$
        & F $\uparrow$ & E $\uparrow$ & S $\uparrow$ & $\mathcal{M}\downarrow$
        & F $\uparrow$ & E $\uparrow$ & S $\uparrow$ & $\mathcal{M}\downarrow$
        \\
        \midrule
        CADC \cite{CADC} & 0.548 & 0.743 & 0.680 & 0.133 & 0.759 & 0.840 & 0.801 & 0.096 & 0.862 & 0.906 & 0.866 & 0.064\\
        GCT-CADC & 0.542 & 0.740 & 0.683 & 0.122 & 0.756 & 0.837 & 0.798 & 0.093 & 0.852 & 0.896 & 0.853 & 0.079\\
        \bottomrule
    \end{tabular}
    \label{tab:performance_on_conventional_cosod_testing_datasets}
\end{table*}

\subsection{Calibration Performance Evaluation}
The proposed GCT also improves the baseline model in terms of the model calibration degree. Fig.~\ref{fig:reliability_diagrams_visualization} illustrates the reliability diagram of CADC and GCT-CADC (Ours). It can be observed that the prediction-to-confidence ratio of GCT-CADC is closer to 1 (Oracle) than the baseline, indicating a higher calibration degree. Further, Tab.~\ref{tab:calibration_comparison} shows the expected calibration error of the existing CoSOD models and our models (GCT-CADC, GCT-DCFM, GCT-ICNet). Each of our models has significantly improvement over the baseline models under both the conventional and generalised CoSOD settings. Furthermore, GCT-CADC achieves the lowest ECE on four out of five testing datasets.

\begin{table}[tbh!]
    \centering
    \renewcommand{\arraystretch}{1.3}
    \renewcommand{\tabcolsep}{0.5mm}
    \caption{CoCA-Common benchmark with different primary ratios.}
    \begin{tabular}{L{0.24\linewidth}|C{0.12\linewidth}C{0.1\linewidth}|C{0.12\linewidth}C{0.1\linewidth}|C{0.12\linewidth}C{0.1\linewidth}}
        \toprule
        {\multirow{2}{*}{Method}} & \multicolumn{2}{|c|}{20 - 40\%} & \multicolumn{2}{c|}{40 - 60\%} & \multicolumn{2}{c}{60 - 80\%} \\
         & mIoU $\uparrow$ &  $\mathcal{M} \downarrow$ & 
        mIoU $\uparrow$ &  $\mathcal{M} \downarrow$ & 
        mIoU $\uparrow$ &  $\mathcal{M} \downarrow$ \\
        \midrule
        EGNet \cite{EGNet} & 44.85 & 0.215 & 49.13 & 0.203 & 53.05 & 0.194\\
        TEP \cite{TEP} & 44.54 & 0.222 & 49.00 & 0.209 & 53.89 & 0.200\\
        \midrule
        ICNet \cite{ICNet}& 46.17 & 0.208 & 52.06 & 0.188 & 57.52 & 0.171\\
        GCAGC \cite{GCAGC}& 63.21 & 0.066 & 61.87 & 0.080 & 61.42 & 0.087\\
        GICD \cite{GICD_CoCA}& 48.74 & 0.156 & 53.49 & 0.152 & 58.10 & 0.141\\
        GCoNet \cite{GCoNet} & 52.65 & 0.084 & 56.32 & 0.096 & 60.89 & 0.101\\
        CADC \cite{CADC}& 48.87 & 0.153 & 54.88 & 0.132 & 59.20 & 0.131\\
        \midrule
        GCT-CADC & \textbf{74.75} & \textbf{0.058} & \textbf{72.06} & \textbf{0.078} & \textbf{69.01} & \textbf{0.086}\\
        \bottomrule
    \end{tabular}
    \label{tab:performance_in_different_primary_group_ratio_range}
\end{table}

\begin{figure*}[ht!]
    \centering
    \includegraphics[width=\textwidth]{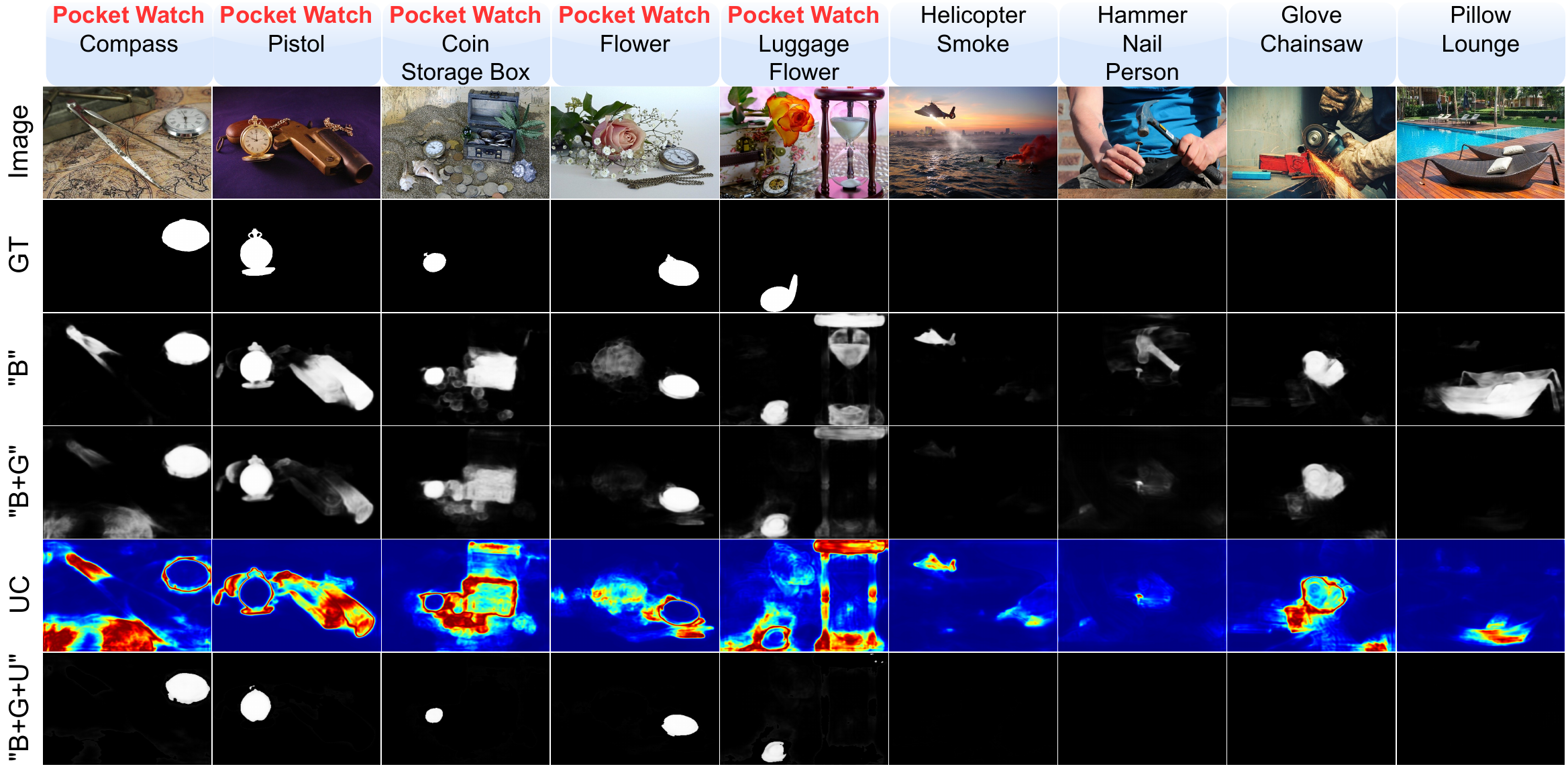}
    \caption{Qualitative results of baseline model (\enquote{B}), Generalised CoSOD Training (\enquote{B+G}) and Uncertainty-based Revision (\enquote{B+G+U}) on the \enquote{Pocket Watch} group of the CoCA-Common dataset. UC is the uncertainty map. After removing the potential false-positive predictions with high uncertainties, the predictions focus more primarily on the common salient object.}
    \label{fig:qualitative_ablation}
\end{figure*}

\subsection{Ablation Study}
\subsubsection{Effectiveness of Generalised CoSOD Training} 
In Tab.~\ref{tab:benchmark_CoSOD_Common_and_Zero}, we observe significantly improved performance compared with the baseline models \cite{CADC}. Take GCT-CADC $vs.$ CADC as an example, it enhances the mIoU scores by 34.2\% and 109.7\%, and reduces the MAE by 48.6\% and 90.1\% on the CoCA-Common and CoCA-Zero datasets respectively. This is attributed to the successful acknowledgement of the absence of common salient object in the \enquote{noisy images}. As a result, GCT-CADC is able to make a full-negative predictions for images where co-salient objects are absent whereas the baseline model tends to output false-positive predictions in \enquote{noisy images}. As shown in Fig.~\ref{fig:qualitative_ablation}, the baseline model (\enquote{B}) falsely identifies the intra-image and inter-image secondary salient objects as co-salient objects. After incorporating the proposed Generalised CoSOD Training (\enquote{B+G}), the false-positive predictions e.g., flower ($4^{th}$ column), helicopter ($6^{th}$ column), lounge ($9^{th}$ column), etc., can be mostly avoided.

\begin{figure*}[htb!]
\centering
\includegraphics[width=\textwidth]{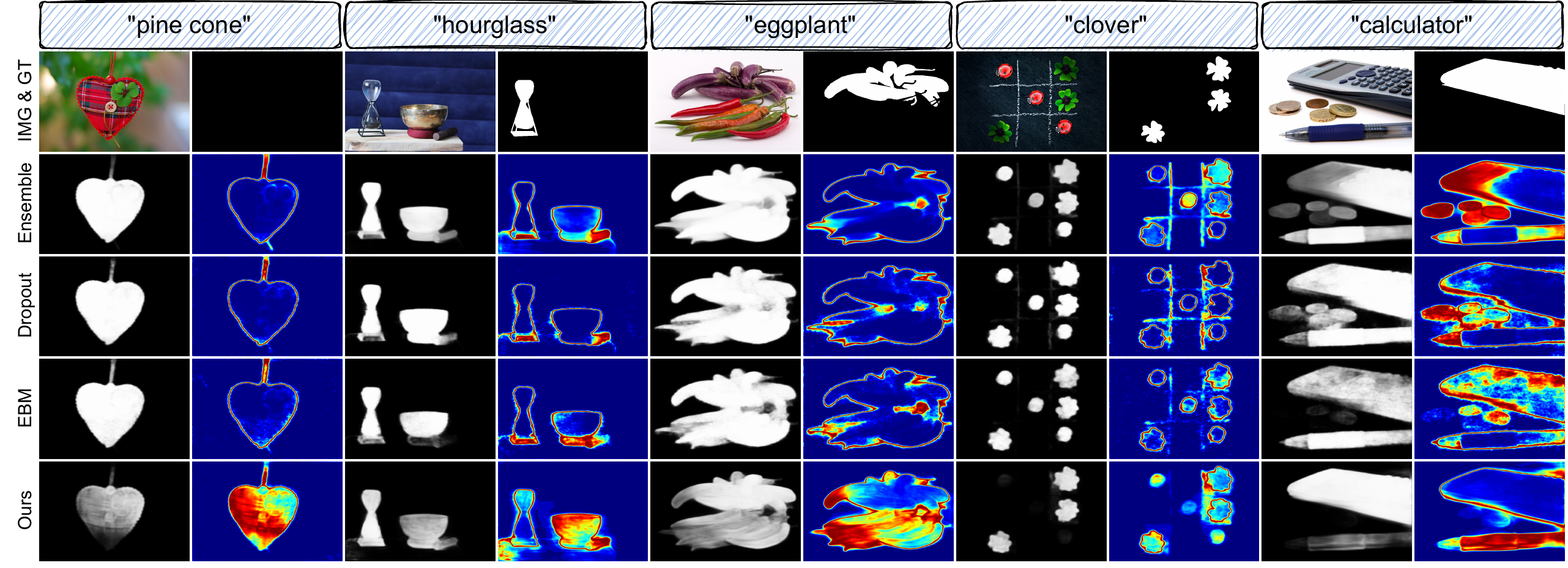}
\caption{Comparison of qualitative results of uncertainty maps estimated with Ensemble, Dropout, Energy Based Model (EBM), and Ours. The header indicates the0 co-salient object of the group, from which the sample is chosen. High uncertainties are represented in \textcolor{red}{red} and low uncertainties in \textcolor{blue}{blue} in the uncertainty maps.}
\label{fig:uncertainty_comparison}
\end{figure*}

\begin{table}[tbh!]
    \centering
    \scriptsize
    \renewcommand{\arraystretch}{1.2}
    \renewcommand{\tabcolsep}{1.3mm}
    \caption{$ECE$ of our GCT and the compared uncertainty estimation methods including Dropout \cite{gal2016dropout}, Ensemble \cite{lakshminarayanan2017simple}, and EBM\cite{zhang2021learning}.}
    \begin{tabular}{l|ccccc}
        \toprule
        Method & CoCA & CoSOD3k  & CoSal2015 & CoCA-Common& CoCA-Zero \\
        \midrule
        Dropout & 0.087 & 0.057 & 0.049 & 0.105 & 0.144\\
        Ensemble & 0.060 & 0.042 & 0.025 & 0.068 & 0.095\\
        EBM & 0.086 & 0.051 & 0.043 & 0.096 & 0.121 \\
        \midrule
        GCT-CADC & \textbf{0.058} & \textbf{0.036} & \textbf{0.016} & \textbf{0.022} & \textbf{0.007}\\ 
        \bottomrule
    \end{tabular}
    \label{tab:uncertainty_comparison}
\end{table}

\subsubsection{Quality of Uncertainty Estimation}
\label{sec:results_effectiveness_of_uncertainty_revision}
The random sampling inherent in GCT enables our model to generate high-quality uncertainty maps. 
As shown in Fig.~\ref{fig:qualitative_ablation}, the uncertainty maps are able to highlight false-positive predictions on secondary salient objects. We further apply a dynamic threshold to the uncertainty map to find potential false-positive co-saliency predictions. Specifically, we compute a bias matrix, measuring the pixel-wise distance between the uncertainty map and its instance-wise mean as $\phi = u(x_{ki}) - \mu_{u_{ki}} \mathbb{A}$, where $\mathbb{A}$ is an all-one matrix, $\mu_{u_{ki}}$ is a single-image uncertainty mean. We remove those unreliable predictions which spatially correspond to positive values in the bias matrix. It can be seen that removing those uncertain predictions further improves the quality of the predictions.

Different from existing uncertainty estimation techniques 
that usually focus on object boundaries \cite{lakshminarayanan2017simple,gal2016dropout,zhang2021learning}, our method estimates the uncertainties at an instance level. Fig.~\ref{fig:uncertainty_comparison} compares the uncertainty estimation qualities of Ensemble \cite{lakshminarayanan2017simple}, Dropout \cite{gal2016dropout}, and Energy-Based Model (EBM) \cite{zhang2021learning} with ours. It can be observed that the compared methods only highlight the boundary of an incorrectly detected non-co-salient object in the noisy image ($1^{\text{st}}$ column). Similarly, for the secondary salient objects that are incorrectly identified as a co-salient object, they also only identify the boundary as uncertain ($2^{\text{nd}}$ - $5^{\text{th}}$ columns). On the contrary, our method can acknowledge the mis-detection in its entirety as uncertainty in both situations. For example, the heart in the \enquote{pine cone} group ($1^{\text{st}}$ column); the chilli pepper in the \enquote{eggplant} group ($3^{\text{rd}}$ column); the pen in the \enquote{calculator} group ($5^{\text{th}}$ column).
Following \cite{ovadia2019can}, we adopt ECE to quantitatively evaluate the estimated uncertainty of our GCT and the compared methods. As shown in Tab.~\ref{tab:calibration_comparison}, our GCT achieves higher ECE scores than the compared methods in both CoSOD and GCoSOD settings, indicating a better reliability of our estimated uncertainty map.



\begin{figure*}[ht!]
    \centering
    \includegraphics[width=\textwidth]{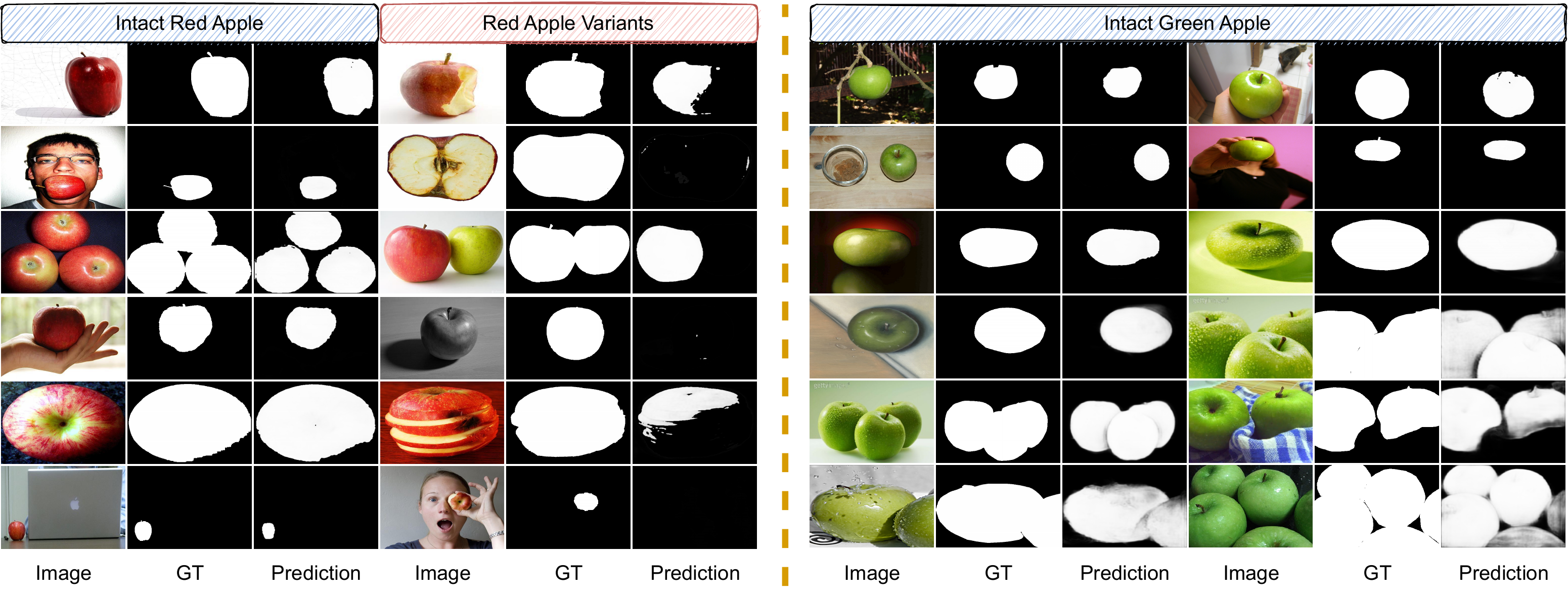}
    \caption{Example images included in the \enquote{apple} group of CoSal2015 \cite{CoSal2015} (left) and CoSOD3k \cite{CoEGNet_CoSOD3k} (right). Intact red apple makes up the majority of the \enquote{apple} group of CoSal2015. This renders our model, which is aware of potential noisy salient objects at inter-image level, to discriminate the intact red apple against its variants including chewed red apple, grayscale apple, peeled apple, and green apple. On the other hand, \enquote{Apple} group of CoSOD3k \cite{CoEGNet_CoSOD3k} contains primarily intact green apple without variants and our model can also acknowledge the universal presence of co-salient object in each individual image.
    }
    \label{fig:false_positive_predictions_in_apple_group}
\end{figure*}

\subsection{Discussion}
\subsubsection{Performance for Conventional CoSOD Datasets}
We further evaluate the proposed \enquote{Generalised CoSOD Training} strategy on conventional CoSOD testing datasets, and show performance in Tab.~\ref{tab:performance_on_conventional_cosod_testing_datasets},using \enquote{CADC} as the baseline model.
It shows that GCT, while significantly enhancing the performance of the baseline model in terms of handling the \enquote{noisy images} in the GCoSOD setting and model calibration degree in both conventional and generalised CoSOD settings, still performs very close to the baseline model under the conventional setting. We also observe improved MAE on CoCA \cite{GICD_CoCA} and CoSOD3k \cite{CoEGNet_CoSOD3k}. The performance drop can be attributed to the awareness of potential absence of common salient object in the \enquote{noisy images}.
Conventional models require a co-salient object to be present in every image, ideally matching the conventional CoSOD datasets.

GCT training allows a model to predict the absence of a co-salient objects in images, which does not occur in conventional CoSOD testing datasets. In this case, GCT-CADC allows discrimination of a common salient object from its different variants, leading to false negative predictions. Fig.~\ref{fig:false_positive_predictions_in_apple_group} illustrates the false negative predictions of the proposed GCT on the apple group of the CoSal2015 dataset \cite{CoSal2015}. The awareness of a potential absence of a co-salient object in individual images of the group makes our model classify the target objects in a more granular scale. Specifically, the intact red apple makes up the majority of the apple group in CoSal2015 \cite{CoSal2015}, resulting in our model discriminating against in various apple variants including \textit{chewed apple}, \textit{sliced apple}, \textit{green apple}, \text{peeled apple}, and \textit{grayscale apple}. On the other hand, the apple group in CoSOD3K \cite{CoEGNet_CoSOD3k} dataset contains intact green apples with consistent appearances and our model successfully detect them as the co-salient object.


\subsubsection{Performance w.r.t. Primary Group Ratios:}
We divide the CoCA-Common dataset into three different groups with primary group ratio in the range of [0.2, 0.4), [0.4, 0.6) and [0.6, 0.8] respectively, and show the performance of existing CoSOD and SOD models on the above individual sets in Tab.~\ref{tab:performance_in_different_primary_group_ratio_range}. We observe decreased performance with the smaller primary group ratio for existing CoSOD and SOD models. This is due to their 
implicit assumption that a co-salient object can be found in each individual image. On the contrary, our method acknowledges the absence of co-salient object, leading to better performance in all ranges of primary group ratio. Tab.~\ref{tab:performance_in_different_primary_group_ratio_range} reveals a converging trend between our method and the baseline model as the primary ratio increases. This is expected as GCT-CADC has better detection of the absence of a co-salient object in \enquote{noisy images} than its baseline. Their performance is similar when the common salient object is present in each individual image of the same group.

\section{Conclusion}
We propose the Generalised Co-Salient Object Detection setting where \enquote{noisy images} that do not share the common salient object can be present in various degrees. We further design a random sampling based Generalised Training strategy, using Diverse Sampling Self-Supervised Learning to distill awareness of the potential absence of the common salient object into existing CoSOD models. In addition, the random sampling process inherent in the GCT allows the model to produce high-quality uncertainty maps that highlight the potential false positive predictions at instance level. Our method is compatible with existing CoSOD models and significantly improves existing CoSOD models in terms of: (1) performance under the GCoSOD setting where \enquote{noisy images} with no common salient object visible can be present in various degrees; and, (2) the model calibration degree under the standard CoSOD setting, preventing over-confident predictions.

\bibliographystyle{IEEEtran}
\bibliography{Reference}

 




\end{document}